\documentclass[smallcondensed]{svjour3}
\usepackage[T1]{fontenc}
\usepackage[utf8]{inputenc}
\usepackage{url}
\usepackage[breaklinks=true]{hyperref}\hypersetup{pdfborder={0 0 0}}
\usepackage{natbib}
\usepackage{amsmath,amssymb}
\usepackage{newtxtext} %
\usepackage{color}
\usepackage{graphicx}
\usepackage{caption}
\usepackage{subcaption}
\usepackage[algo2e,noend,linesnumbered]{algorithm2e}
\makeatletter\renewcommand{\@algocf@capt@plain}{above}\makeatother
\usepackage{tikz}

\newcommand{\TD}{\ensuremath{\mathit{TD}}}
\newcommand{\SSQ}{\ensuremath{\mathit{SSQ}}}
\newcommand{\dist}{\ensuremath{\mathit{d}}}
\DeclareMathOperator*{\argmax}{arg\,max}
\DeclareMathOperator*{\argmin}{arg\,min}
\newcommand{\kmedoids}{\mbox{$k$-medoids}}
\newcommand{\change}{\ensuremath{\Delta}}
\newcommand{\best}[1]{{#1\mathstrut}^*\!}
\newcommand{\Change}{\Delta\TD}

\newcommand{\reffig}[1]{Figure~\ref{#1}}
\newcommand{\refsec}[1]{Section~\ref{#1}}
\newcommand{\refalg}[1]{Algorithm~\ref{#1}}
\newcommand{\refeqn}[1]{Equation~\ref{#1}}

\title{Faster \boldmath$k$-Medoids Clustering:\\
Improving the PAM, CLARA, and CLARANS Algorithms}
\titlerunning{Faster $k$-Medoids Clustering: Improving PAM, CLARA, and CLARANS}
\author{Erich Schubert \and Peter J.{} Rousseeuw}
\institute{
Erich Schubert \at
Technische Universität Dortmund, Dortmund, Germany\\
\email{erich.schubert@tu-dortmund.de}
\and
Peter J.{} Rousseeuw \at
Department of Mathematics, KU Leuven, Leuven, Belgium\\
\email{peter@rousseeuw.net}
}

\date{\vspace{-2cm}\\
\textcolor{red}{
\textbf{Preprint} dated 2019-05-04. Please consult the final version instead:
\\
Erich Schubert, Peter J. Rousseeuw:
Faster k-Medoids Clustering: Improving the PAM, CLARA, and CLARANS Algorithms.
Similarity Search and Applications. SISAP 2019: 171-187
\url{https://doi.org/10.1007/978-3-030-32047-8_16}
}
}
\begin{document}

\maketitle

\begin{abstract}
Clustering non-Euclidean data is difficult, and one of the most used
algorithms besides hierarchical clustering is the popular algorithm
Partitioning Around Medoids (PAM), also simply referred to as $k$-medoids.

In Euclidean geometry the mean---as used in $k$-means---is a good estimator for the cluster
center, but this does not exist for arbitrary dissimilarities. PAM uses the
medoid instead, the object with the smallest dissimilarity to all others
in the cluster. This notion of centrality can be
used with any (dis-)similarity, and thus is of high relevance to many
domains such as biology that require the use of
Jaccard, Gower, or more complex distances.

A key issue with PAM is, however, its high run time cost.
In this paper, we propose modifications to the PAM algorithm
where at the cost of storing $O(k)$ additional values,
we can achieve an $O(k)$-fold speedup in the second (``SWAP'') phase
of the algorithm, but will still find the same results as the original PAM algorithm.
If we slightly relax the choice of swaps performed (while retaining comparable
quality), we can further accelerate the algorithm by performing up to
$k$ swaps in each iteration.
With the substantially faster SWAP, we can now also explore alternative (faster)
strategies for choosing the initial medoids.
We also show how the CLARA and CLARANS
algorithms benefit from the proposed modifications.

While we do not further study the parallelization of our approach in this work, it can easily
be combined with earlier approaches to use PAM and CLARA on big data (some of
which use PAM as a subroutine, hence can immediately benefit from these improvements),
where the performance with high $k$ becomes increasingly important.

In experiments on real data with $k=100$, we observed a $200\times$ speedup
compared to the original PAM SWAP algorithm, making PAM applicable to larger data
sets as long as we can afford to compute a distance matrix, and in particular to higher~$k$
(at $k=2$, the new SWAP was only 1.5 times faster, as the speedup is expected to increase with $k$).

\end{abstract}

\section{Introduction}

Clustering is a common unsupervised machine learning task, in which the data set
has to be automatically partitioned into ``clusters'', such that objects within
the same cluster are more similar, while objects in different clusters are
more different.
There is not (and likely never will be) a generally accepted definition
of a cluster, because ``clusters are, in large part, in the eye of the
beholder''~\citep{DBLP:journals/sigkdd/Estivill-Castro02}, meaning that
every user may have different enough needs and intentions to want a
different algorithm and notion of cluster.
And therefore, over many years of research,
hundreds of clustering algorithms
and evaluation measures have been proposed, each with their merits and drawbacks.
Nevertheless, a few seminal methods such as hierarchical clustering, $k$-means,
PAM~\citep[][Ch.{} 2]{KauRou87,KauRou90}, and \mbox{DBSCAN}~\citep{DBLP:conf/kdd/EsterKSX96}
have received repeated and widespread use. One may be tempted
to think that after 60 to 20 years these methods have all been well researched and
understood, but there are still many scientific publications 
trying to explain these algorithms better (e.g.,~\citealt{DBLP:journals/tods/SchubertSEKX17}),
trying to parallelize and scale them to larger data sets (e.g.,~\citealt{DBLP:journals/datamine/LijffijtPP15,serss/YangL14}),
trying to better understand similarities and relationships among the published methods
(e.g.,~\citealt{DBLP:conf/lwa/SchubertHM18}),
or proposing further improvements --
and so does this paper for the widely used PAM algorithm, also often referred to as \kmedoids{}. 

In hierarchical agglomerative clustering (HAC), each object is initially its
own cluster. The two closest clusters are then merged repeatedly to build a cluster
tree called dendrogram. HAC is a very flexible method: it can be used with any
distance or (dis-)similarity, and it allows for different rules of aggregating
the object distances into cluster distances, such as the minimum (``single linkage''),
average, or maximum (``complete linkage''). Single linkage directly
corresponds to the minimum spanning tree of the distance graph.
While the dendrogram is a powerful visualization for small data sets, extracting flat
partitions from hierarchical clustering is not trivial, and thus users
often turn to simpler methods.

Another classic method taught in textbooks is $k$-means (for an overview of
the complicated history of $k$-means, refer to \citealt{Bock07}), where the data is
modeled using $k$ cluster means, that are iteratively refined by assigning all objects
to the nearest mean, then recomputing the mean of each cluster.
This optimization converges because the mean is the least squares estimator of location,
and both steps optimize the same quantity, a measure known as sum-of-squared errors~(\SSQ, also called SSE, and equivalent to WCSS):
\begin{align}
\SSQ :=&
\sum\nolimits_{i=1}^k \sum\nolimits_{x_j \in C_i} ||x_j - \mu_i||_2^2
\,.
\label{eqn:ssq}
\end{align}

In \kmedoids{}, the data is modeled very similarly, but using $k$ representative objects $m_i$
called medoids (chosen from the data set; defined below) that can serve as
``prototypes'' for the cluster
instead of means in order to allow using arbitrary other dissimilarities and
arbitrary input domains,
using the absolute error criterion (``total deviation'', \TD) as objective:
\begin{align}
\TD :=&
\sum\nolimits_{i=1}^k \sum\nolimits_{x_j \in C_i} \dist(x_j, m_i)
\,,
\label{eqn:td}
\end{align}
which is the sum of dissimilarities of each point $x_j\in C_i$ to the medoid~$m_i$ of its cluster.
If we use squared Euclidean as distance function (i.e., $d(x,m)=||x-m||_2^2$),
we almost obtain the usual \SSQ{} objective used by $k$-means, except that $k$-means is free to choose any
$\mu_i\in\mathbb{R}^d$, whereas in \kmedoids{} $m_i\in C_i$ must be one of the original
data points.
For \emph{squared} Euclidean distances and Bregman divergences,
the arithmetic mean is the optimal choice for $\mu$ and a fixed cluster assignment.
For $L_1$ distance (i.e, $\sum |x_i-y_i|$), also called Manhattan distance,
the component-wise median is a better choice in $\mathbb{R}^d$
~\cite[$k$-medians]{DBLP:conf/nips/BradleyMS96}. For unsquared Euclidean distances,%
\footnote{It is a common misconception that $k$-means would minimize Euclidean distances.
It optimizes the sum of \emph{squared} Euclidean distances, and even then the textbook
algorithm may end up slightly off a local optimum, because always assigning a point to its nearest center \emph{can}
increase the variance by moving the centers away from other points \citep{HartiganW79}.}
we get the much harder Weber problem~\citep{DBLP:journals/mp/Overton83},
which has no closed-form solution~\citep{DBLP:conf/nips/BradleyMS96}
(for a recent survey of algorithms for the Weber point see \citealt{FritzFC12}).
For other distance functions, finding a closed form to compute the best $m_i$ would require
non-trivial mathematical analysis of each distance function separately. Furthermore, our input
data does not necessarily come from a $\mathbb{R}^d$ vector space.
In \kmedoids{} clustering, we therefore constrain $m_i$ to be one of our data samples.
The medoid of a set $C$ is defined as the object with the smallest sum of dissimilarities
(or, equivalently, smallest average)
to all other objects in the set:
$$
\operatorname{medoid}(C) := \argmin_{x_i\in C} \sum\nolimits_{x_j\in C} \dist(x_i, x_j)
$$
This definition does not require the dissimilarity to be a metric,
and by using $\argmax$ instead of the $\argmin$ it can also be applied to similarities.
The algorithms discussed below all can trivially be modified to maximize similarities
rather than minimizing distances, and none assumes the triangular inequality.
Partitioning Around Medoids (PAM, \citealt[][Ch.{} 2]{KauRou87,KauRou90}) is the most widely known algorithm to
find a good partitioning using medoids, with respect to $\TD$ (\refeqn{eqn:td}).

\section{Partitioning Around Medoids (PAM)}

The ``Program PAM''~\citep[][Ch.{} 2]{KauRou87,KauRou90} consists of two algorithms,
BUILD to choose an initial clustering, and SWAP
to further improve the clustering towards a local optimum (finding
the global optimum of the $k$-medoids problem is, unfortunately, NP-hard). 
The algorithms require a dissimilarity matrix
(for example computed using the routine DAISY of \citealt{KauRou90}),
which requires $O(n^2)$ memory and
for many popular distance functions in $d$ dimensional data
$O(n^2 d)$ time to compute (but potentially much more for expensive distances such as
earth movers distance).
Computing the distance matrix will therefore in many cases be much of the computational cost already.

\subsection{BUILD Initialization Algorithm}

\begin{algorithm2e}[t]
\caption{PAM BUILD: Find initial cluster centers.}
\label{alg:build}
$(\TD,m_1) \leftarrow (\infty, \text{null})$\;
\ForEach(\tcp*[f]{First medoid}){$x_j$}{
  $\TD_j\leftarrow 0$ \;
  \lForEach{$x_o \neq x_j$}{
    $\TD_j \leftarrow \TD_j + \dist(x_o,x_j)$%
  }
  \lIf(\tcp*[f]{Smallest distance sum}){$\TD_j<\TD$}{
    $(\TD, m_1) \leftarrow (\TD_j, x_j)$%
  }
}
\For(\tcp*[f]{Other medoids}){$i=1\ldots k-1$}{
  $(\best{\Change}, \best{x}) \leftarrow (\infty, \text{null})$\;
  \ForEach{$x_j \not\in \{m_1,\ldots,m_i\}$}{
    $\Change\leftarrow 0$ \;
    \ForEach{$x_o \not\in \{m_1,\ldots,m_i,x_j\}$}{
      $\delta \leftarrow \dist(x_o,x_j) - \min_{o \in m_1,\ldots,m_i}\dist(x_o, o)$\;
      \label{line:build-cached}
      \lIf{$\delta<0$}{$\Change\leftarrow\Change + \delta$}
    }
    \lIf(\tcp*[f]{best reduction in TD}){$\Change<\best{\Change}$}{
      $(\best{\Change}, \best{x}) \leftarrow (\Change, x_j)$%
    }
  }
  $(\TD, m_{i+1}) \leftarrow (\TD + \best{\Change}, \best{x})$\;
}
\Return $\TD, \{m_1,\ldots,m_k\}$\;
\end{algorithm2e}

In order to find a good initial clustering
(rather than relying on a random sampling
strategy as commonly used with $k$-means),
BUILD chooses $k$~times the point which yields the smallest distance sum \TD{}
(in the first iteration, this means choosing the point with the smallest distance
to all others; afterwards adding the point as next medoid, that reduces \TD{} most).
We give a pseudocode in \refalg{alg:build}, where we use $\Change$ as symbol for the
change in $\TD$ (which should be negative to be beneficial), and $\best{}$ for the best values found so far.
If we cache the nearest medoid in line~\ref{line:build-cached}, then
BUILD initialization needs $O(n^2 k)$ time, so it already is a fairly expensive algorithm.
The motivation here was to find a good starting point, in order to require fewer
iterations of the refinement procedure.
In the experiments, we will also study
whether a clever sampling-based approach similar to k-means++~\citep{DBLP:conf/soda/ArthurV07}
that needs only $O(nk)$ time but produces a worse starting point
is an interesting alternative.

\subsection{SWAP Refinement Algorithm}

\begin{algorithm2e}[t]
\caption{PAM SWAP: Iterative improvement.}
\label{alg:swap}
\SetKwFor{Repeat}{repeat}{}{end}
\SetKw{StopIf}{break loop if}
\Repeat{}{
  $(\best{\Change}, \best{m}, \best{x}) \leftarrow (0, \text{null}, \text{null})$ \;
  \ForEach(\tcp*[f]{each medoid}){$m_i \in \{m_1,\ldots,m_k\}$}{
    \ForEach(\tcp*[f]{each non-medoid}){$x_j \not\in \{m_1,\ldots,m_k\}$}{
      $\Change\leftarrow 0$ \;
      \lForEach{$x_o\not\in \{m_1,\ldots,m_k\}\setminus m_i$}{
        $\Change \leftarrow\Change + \change(x_o,m_i,x_j)$ 
        \label{line:swap-change}
      }
      \lIf{$\Change < \best{\Change}$}{
        $(\best{\Change}, \best{m}, \best{x}) \leftarrow (\Change, m_i, x_j) $
      }
    }
  }
  \StopIf $\best{\Change}\geq 0$ \;
  swap roles of medoid $\best{m}$ and non-medoid $\best{x}$ \tcp*{perform best swap}\label{line:swap-swap}
  $\TD \leftarrow \TD + \best{\Change}$ \;
}
\Return $\TD,M,C$\;
\end{algorithm2e}

The second algorithm, which is the main focus of this paper, was named SWAP.
In~order to improve the clustering, it considers all possible changes to the set of $k$~medoids,
which effectively means replacing (swapping) some medoid with some non-medoid, which gives
$k(n-k)$ candidate swaps.
If it reduces \TD{},
the best such change is then applied, in the spirit of a greedy steepest-descent method,
and this process is repeated until no further improvements are found.
We give a pseudocode of this in \refalg{alg:swap}.
If we again cache the necessary data to compute the $\change(x_o,m_i,x_j)$ function
(\refeqn{eqn:change}, explained in \refsec{sec:best-swap})
in line~\ref{line:swap-change} efficiently, then the run time of this algorithm is $O(k (n-k)^2)$ for each iteration.
While the authors of PAM assumed that only few iterations will be needed (if the algorithm
is already initialized well, using the BUILD algorithm above),
we do see an increasing number of iterations
with increasing amounts of data (but usually we will have fewer than $k$ iterations).

Both the pseudocode for BUILD and SWAP given here omit the details of managing the cached distances.
For each object, we need to store the index of the nearest medoid $\textit{nearest}(o)$
and its distance $\dist_{\textit{nearest}}(o)$, and also the distance to the second nearest medoid
$\dist_{\textit{second}}(o)$ (if we also store the index of the second nearest center,
we may be able to avoid some more distance computations). In particular in line~\ref{line:swap-swap},
when executing the best swap, we need to carefully update the cached values.

\subsection{Variants of PAM}

The algorithm CLARA~(\citealt{KaufmanR86}, \citeyear[][Ch.{} 3]{KauRou90}) repeatedly applies PAM on a subsample with
$n^\prime\ll n$ objects, with the suggested value $n^\prime=40+2k$.
Afterwards, the remaining objects are assigned to their closest
medoid. The run with the least $\TD$ (on the entire data) is returned. If the sample size
is chosen $n^\prime\in O(k)$ as suggested, the run time reduces to $O(k^3)$, which explains why the approach
is typically used only with small $k$~\citep{LucasiusDK93}. Because CLARA uses PAM internally,
it will directly benefit from the improvements proposed in this article.

\citet{LucasiusDK93} propose a genetic algorithm to find the best \kmedoids{} partitioning,
which will perform a randomized exploration of the search space based on ``mutation'' of the best solutions
found so far. Crossover mutations correspond to taking some medoids from both ``parents'',
whereas mutations replace medoids with random objects.
It is not obvious that this will efficiently provide a sufficient coverage of the enormous
search space (there are $\tbinom{n}{k}=\frac{n!}{k!(n-k)!}$ possible sets of medoids) for a large $k$.
In order to benefit from the proposed improvements, a more systematic mutation strategy would
need to be adopted, making the method similar to CLARANS below.
\citet{DBLP:journals/eswa/WeiLH03} found the genetic methods to work
only for small data sets, small $k$, and well separated symmetric clusters,
and it was usually outperformed by CLARANS.

The algorithm CLARANS~\citep{NgHan02} interprets the search space as a high-dimensional hypergraph,
where each edge corresponds to swapping a medoid and non-medoid. On this graph it performs a
randomized greedy exploration, where the first edge that reduces the loss $\TD$ is followed
until no edge can be found with $p= 1.25\% \cdot k(n-k)$ attempts.
In \refsec{sec:better-clarans} we will outline how our approach can be used to explore $k$
edges at a time efficiently; this will allow exploring a larger part of the search space in similar
time, but we expect the savings to be relatively small compared to PAM.

\citet{DBLP:journals/jmma/ReynoldsRIR06} discuss an interesting trick to speed
up PAM. They show how to decompose the change in the loss function into two components,
where the first depends only on the medoid removed, the second part only on the new point.
This decomposition forms the base for our approach, and we will thus discuss it in
\refsec{sec:best-swap} in more detail. 

\citet{DBLP:journals/eswa/ParkJ09} propose a ``$k$-means like'' algorithm for \kmedoids{}
(which actually was already considered by \citealt{DBLP:journals/jmma/ReynoldsRIR06} before),
where in each iteration the medoid is chosen to be the object with the smallest distance sum to
other members of the cluster, then each point is assigned to the nearest medoid until
\TD{} no longer decreases.
This is, unfortunately, not very effective at improving the clustering:
new medoids are only chosen from within the cluster, and \emph{have} to cover the entire current
cluster. This misses many improvements where cluster members can be reassigned to \emph{other} clusters
with little cost; such improvements are, however, found by PAM. Furthermore, the means used in $k$-means
change with every point we move to a different cluster, but the medoids will very often remain the same,
they are too coarse for this optimization strategy.
In our experiments this approach produced noticeably worse results than PAM, in line with
the earlier observations by \citet{DBLP:journals/jmma/ReynoldsRIR06}.
The paper also contributes an $O(n^2)$ initialization, that unfortunately tends to choose all
initial medoids close to the center of the data set.
Choosing cluster medoids with a $k$-means like strategy will take $O(n^2)$ time because we have to
assume the clusters to be unbalanced, and contain up to $O(n)$ objects;
nevertheless this is $k$ times faster than PAM.
But because it does \emph{not} consider reassigning points to other clusters when choosing the new
medoid, this approach will miss many improvements found by PAM. This reduces
the number of iterations, but also produces worse results.

\citet{DBLP:journals/datamine/LijffijtPP15} propose to use the
$k$-means++ \citep{DBLP:conf/soda/ArthurV07} initialization
with PAM and CLARA, but give the resulting complexity incorrectly,
missing a factor of $k$. They do not study the effect of replacing the initialization,
which requires more iterations to converge (and hence, will be slower with original PAM),
as we will discuss later in \refsec{sec:fasterinit}, and demonstrate in \refsec{sec:numiter}.

Since PAM needs $O(n^2)$ memory for the distance matrix, it is not usable on big data.
Therefore, people have proposed various approximations to PAM, such as CLARA and CLARANS
discussed before.
\citet{serss/YangL14} parallelize the ``$k$-means like'' variant with map-reduce,
parallelizing over the cluster in the reduce step. When cluster sizes vary substantially,
this needs $O(n^2)$ memory in the reducer, and may yield next to no speedup in the worst case.
CLARA can be trivially parallelized by randomly partitioning the data,
then running PAM on each partition \citep{journals/compstat/KaufmanLR88}. %
This approach will obviously benefit from our improvements the same way as CLARA and PAM benefit.
A recent example is PAMAE~\citep{DBLP:conf/kdd/Song0H17}, which essentially is
CLARA with an additional refinement step: it draws random samples and runs
any $k$-medoids approach on each; chooses the best medoids found, and refines them
with a single iteration of a approximate parallel version of the ``$k$-means like''
update.
Papers have rarely considered using large $k$ values,
although this makes sense in the context
of data approximation, where you want to reduce the data set to $k$ representative samples.
Many of the attempts at distributing and parallelizing PAM employ
PAM as a subroutine, and hence can trivially integrate our improvements.

\section{Finding the Best Swap}\label{sec:best-swap}

We focus on improving the original PAM algorithm here, which is a commonly
used subroutine even in the faster variants such as CLARA (which uses PAM
as a subroutine, and hence directly benefits from any improvement to PAM).
We also discuss how we can obtain similar improvements for CLARANS in \refsec{sec:better-clarans}.

The algorithm SWAP evaluates every swap of each medoid $m_i$ with any non-medoid~$x_j$.
Recomputing the resulting $\TD$ using \refeqn{eqn:td} every time requires finding the nearest medoid
for every point, which causes many redundant computations.
Instead, PAM only computes the \emph{change} in $\TD$ for each object $x_o$
if we swap $m_i$ with $x_j$ separately:
\begin{align}
\Change =&
\textstyle\sum\nolimits_{x_o} \change(x_o,m_i,x_j)
\end{align}
In the function $\change(x_o,m_i,x_j)$ we can often detect when a point remains
assigned to its current medoid (if $c_k\neq c_i$, and this distance is also smaller than the distance to $x_j$),
and then immediately return $0$.
Because of space restrictions, we do not repeat the original ``if'' statements
used in~\citep[][Ch.{} 2]{KauRou90}, but instead condense them directly into the following equation:
\begin{align}
\change(x_o,m_i,x_j)\!=\!&
\begin{cases}
\min\{\dist(x_o,x_j),\dist_s(o)\}-\dist_n(o)
&%
\text{if }i=\textit{nearest}(o)
\\
\min\{
\dist(x_o,x_j)
-\dist_n(o), 0\}
&%
\text{otherwise}
\end{cases}
\label{eqn:change}
\end{align}
where $\dist_n(o)$ is the distance to the nearest medoid of $o$,
and $\dist_s(o)$ is the distance to the second nearest medoid.
Computing them on the fly would increase the cost of SWAP by
a factor of $O(k)$, but we can cache these values, and only update
them when performing a swap.

\citet{DBLP:journals/jmma/ReynoldsRIR06} note that we can decompose $\Change$ into:
(i)~the loss of removing medoid $m_i$, and assigning all of its cluster members to the next best alternative,
which can be computed as $\sum_{o\in C_i}\dist_s(o)-\dist_n(o)$
(ii)~the (negative) loss of adding the replacement medoid $x_j$, and reassigning all objects closest
to this new medoid.
Since (i) does not depend on the choice of $x_j$, we can make the loop over all medoids $m_i$ outermost,
reassign all its points to the second nearest medoid (cache the distance to the now nearest neighbor),
and compute the resulting loss. We then iterate over all non-medoids and compute the benefit of
using them as the missing medoid instead. In the $\change$ function, we no longer have to consider
the second nearest now (we virtually removed the old medoid already).
The authors observed roughly a two-fold speedup using this approach,
and so do we in our experiments.

Our approach is based on a similar idea of exploiting redundancy in these computations (by caching shared computations),
but we instead move the loops over the medoids~$m_i$ into the \emph{innermost} for loop.
The reason for this is to further remove redundant computations. This becomes apparent when we realize that
in \refeqn{eqn:change}, the second case does not depend on the current medoid $i$.
If we transform the second case into an if statement, we can often avoid to iterate over all $k$ medoids.

\subsection{Making PAM SWAP faster: FastPAM1}

\begin{algorithm2e*}[t!]
\caption{FastPAM1: Improved SWAP algorithm}
\label{alg:iswap}
\SetKwFor{Repeat}{repeat}{}{end}
\SetKw{StopIf}{break loop if}
\Repeat{}{
  $(\best{\Change}, \best{m}, \best{x}) \leftarrow (0, \text{null}, \text{null})$ \tcp*{Empty best candidate storage}\label{line:best}
  \ForEach{$x_j \not\in \{m_1,\ldots,m_k\}$}{
    $d_j\leftarrow \dist_{\text{nearest}}(x_j)$ \tcp*{Distance to current medoid}\label{line:init}
    $\Change\leftarrow (-d_j, -d_j, \ldots, -d_j)$ \tcp*{Loss change for making j a medoid}\label{line:init2}
    \ForEach{$x_o\neq x_j$}{
      $d_{oj}\leftarrow \dist(x_o,x_j)$ \tcp*{Distance to new medoid}
      $(n, d_n, ds) \leftarrow (\text{nearest}(o), \dist_{\text{nearest}}(o), \dist_{\text{second}}(o))$ \tcp*{Cached values}
      $\Change_n \leftarrow \Change_n + \min\{d_{oj},d_s\}-d_n$ \tcp*{Loss change for current}\label{line:casea}
      \If(\tcp*[f]{Reassignment check}){$d_{oj}<d_n$}{\label{line:caseb}\label{line:skip}
      \ForEach{$m_i \in \{m_1,\ldots,m_k\}\setminus m_n$}{
        $\Change_i \leftarrow\Change_i + d_{oj}-d_n$\tcp*{Update loss change}\label{line:skip2}
      }}
    }
    $i\leftarrow \argmin \Change_i$ \tcp*{Choose best medoid i}
    \lIf(\tcp*[f]{Remember best}){$\Change_i < \best{\Change}$}{
      $(\best{\Change}, \best{m}, \best{x}) \leftarrow (\Change_i, m_i, x_j) $
    }
  }
  \StopIf $\best{\Change}\geq 0$\;
  swap roles of medoid $\best{m}$ and non-medoid $\best{x}$ \;
  $\TD \leftarrow \TD + \best{\Change}$ \;
}
\Return $\TD,M,C$\;
\end{algorithm2e*}

\refalg{alg:iswap} shows the improved SWAP algorithm.
In lines~\ref{line:init}--\ref{line:init2}
we compute the benefit of making $x_j$ a medoid. As we do not yet decide which medoid to remove,
we use an array of $\Change$ for each possible medoid to replace. We can now for each point
compute the benefit when removing its current medoid (line~\ref{line:casea}), or the benefit if the
new medoid is closer than the current medoid (line~\ref{line:caseb}), which corresponds to the
two cases in \refeqn{eqn:change}. The interesting property is now since the second case does not
depend on $i$, we can replace the $\min$ statement with an if conditional \emph{outside} of the
loop in lines~\ref{line:skip}--\ref{line:skip2}. After iterating over all points, we choose the
best medoid, and remember the overall best swap. If we always prefer the smaller index $i$ on ties,
we choose \emph{exactly the same} swap as the original PAM algorithm.

\subsection{Benefits and Costs}

If we assume that the new medoid is closest in $O(1/k)$ cases on average (this assumes a somewhat
balanced cluster size distribution),
then we can compute the change for all $k$ medoids with $O(k\cdot 1/k)=O(1)$
effort, by saving the innermost loop in line~\ref{line:skip2}.
Therefore, we expect a typical speedup on the order of $O(k)$
compared to the original PAM SWAP (but it may be hard to guarantee this
for any useful assumption on the data distribution; the worst case supposedly remains unaffected)
at the slight cost of
temporarily storing one $\Change$ for each medoid $m_i$ (compared to the cost of storing the
distance matrix and the distances to the nearest and second nearest medoids,
the cost of this is negligible).

\subsection{Swapping Multiple Medoids: FastPAM2}

\begin{algorithm2e*}[t!]
\caption{FastPAM2: SWAP with multiple candidates}
\label{alg:i2swap}
\SetKwFor{Repeat}{repeat}{}{end}
\SetKw{StopIf}{break loop if}
\SetKw{KwAnd}{and}
\SetKw{Where}{where}
\Repeat{}{
  \lForEach{$x_o$}{
  	compute $\text{nearest}(o), \dist_{\text{nearest}}(o), \dist_{\text{second}}(o)$
  }
  $\best{\Change}, \best{x} \leftarrow [0,\ldots,0], [\text{null}, \ldots, \text{null}]$ \tcp*{Empty best candidates array}
  \label{line:vecstorage}
  \ForEach{$x_j \not\in \{m_1,\ldots,m_k\}$}{
    $d_j\leftarrow \dist_{\text{nearest}}(x_j)$ \tcp*{Distance to current medoid}
    $\Change\leftarrow (-d_j, -d_j, \ldots, -d_j)$ \tcp*{Loss change for making j a medoid}
    \ForEach{$x_o\neq x_j$}{
      $d_{oj}\leftarrow \dist(x_o,x_j)$ \tcp*{Distance to new medoid}
      $(n, d_n, ds) \leftarrow (\text{nearest}(o), \dist_{\text{nearest}}(o), \dist_{\text{second}}(o))$ \tcp*{Cached}
      $\Change_n \leftarrow \Change_n + \min\{d_{oj},d_s\}-d_n$ \tcp*{Loss change for current}
      \If(\tcp*[f]{Reassignment check}){$d_{oj}<d_n$}{
      \ForEach{$m_i \in \{m_1,\ldots,m_k\}\setminus m_n$}{
        $\Change_i \leftarrow\Change_i + d_{oj}-d_n$\tcp*{Update loss change}
      }}
    }
    \ForEach{$i$ \Where $\Change_i < \best{\Change}_i$}{
      $(\best{\Change}_i, \best{x}_i) \leftarrow (\Change_i, x_j)$ \tcp*{Remember the best swap for i}
      \label{line:allbest}
    }
  }
  \StopIf $\min\best{\Change}\geq 0$ \tcp*{At least one improvement was found}
  \While(\tcp*[f]{Execute all improvements}){$i\leftarrow \argmin \best{\Change}$ \KwAnd $\best{\Change}_i < 0$}{
    swap roles of medoid $m_i$ and non-medoid $\best{x}_i$ \;
    \label{line:i2best}
    $\TD \leftarrow \TD + \best{\Change}_i$ \;
    $\best{\Change}_i \leftarrow 0$ \tcp*{Prevent further processing}
    \ForEach(\tcp*[f]{Recompute TD for remaining swaps}){$j$ \Where $\best{\Change}_j<0$}{
      $\Change\leftarrow 0$ \;
      \lForEach{$x_o\not\in \{m_1,\ldots,m_k\}\setminus m_j$}{
        $\Change \leftarrow \Change + \change(x_o,m_j,\best{x}_j)$ 
      }
      \label{line:i2update}
      \lIf(\tcp*[f]{Tolerance check}){$\Change\leq \tau\cdot \best{\Change}_j$}{
        $\best{\Change}_j \leftarrow \Change$
      }
      \label{line:i2tolerance}
      \lElse(\tcp*[f]{Skip otherwise}){$\best{\Change}_j \leftarrow 0$}
    }
  }
}
\Return $\TD,M,C$\;
\end{algorithm2e*}

A second technique to make this second stage of PAM faster is based on the following observation:
PAM will always identify the \emph{single} best swap, then restart search; whereas the classic
$k$-means reassigns all points, and updates all means in each iteration.
Choosing the best swap has the benefit that this makes the algorithm independent of the data order
\citep[][Ch.{} 2]{KauRou90} as long as there are no ties,
and it also means we need to execute fewer swaps than if we would greedily
perform any swap that yields an improvement
(where we may end up replacing the same medoid several times). 

But on the other hand, in particular for large $k$, we can assume that many clusters will be
independent, and we could therefore update the medoids of these clusters in the same iteration.
Naively assuming that each medoid would be swapped in each iteration, this would allow us to reduce
the number of iterations by $k$.

Based on this observation, we propose to consider the best swap for \emph{each} medoid,
and not only the single best swap, i.e., perform up to $k$ swaps.
This is a fairly simple modification shown in \refalg{alg:i2swap}, as we can
simply store an array of swap candidates $(\best{\Change}_i, \best{x}_i)$ in line~\ref{line:vecstorage},
storing one best candidate for each current medoid~$m_i$, and update these in line~\ref{line:allbest}.
After evaluating all possible swaps, we find the best swap within these up to $k$~candidates
(if we did not find a candidate, the algorithm has converged).
We perform the best of these swaps in line~\ref{line:i2best}, mark it as done.
Then we have to recompute in line~\ref{line:i2update} for each remaining swap candidate
if it still improves the clustering, otherwise the swap is not performed.
At this point, in line~\ref{line:i2tolerance}, we explore two alternatives:
(a)~``strict'' ($\tau=1$): only swaps are executed that appear to be independent of
the previous swaps (because their $\Change$ has not changed)
and (b)~``greedy'' ($\tau=0$): execute any swap that still yields a benefit.

The benefits of this strategy are, unsurprisingly, much smaller than the first improvement. In early
iterations we see multiple swaps being executed, but in the later iterations it is common that only
few medoids change at all. Nevertheless, this simple modification does yield another measurable
performance improvement. However---in contrast to the first improvement---this no longer guarantees to
yield the exact same result (because the additional swaps may occasionally be worse than the swaps
found when rescanning, or may be executed in a slightly different order).
From a theoretical point of view, both the original PAM, and FastPAM2 perform a
steepest descent optimization strategy; where PAM only permits descends consisting of a single swap,
whereas FastPAM2 can perform multiple swaps at once as long as they use different medoids.
Therefore, we argue that both are able to find results of equivalent quality.\footnote{Neither
can guarantee to find the global optimum, which would be NP-hard.}
In our experiments, even the ``greedy'' strategy would often find slightly better results
than PAM, and faster. %

\subsection{Faster Initialization with Linear Approximative BUILD (LAB): FastPAM}\label{sec:fasterinit}

With these optimizations to the SWAP algorithm,
that reduce the run time from $O(k(n-k)^2)$ to $O((n-k)^2)$,
the main bottleneck of PAM suddenly becomes the first algorithm, BUILD.
In the experiments below on the plant species data at $k=200$,
using the R implementation, PAM would spend 99\% of the run time in SWAP.
With above optimizations this reduces to about 15\%. About 16\% is the time to compute the distance
matrix, and 69\% of the time is spent in BUILD. The run time of BUILD is in $O(kn^2)$,
so for large $k$ this is not unexpected to happen.
But since we were able to make SWAP much faster, we can now afford to begin with
slightly worse starting conditions, even if we need more iterations of SWAP afterwards.

A very elegant way of choosing starting conditions in $k$-means is known as
$k$-means++ \citep{DBLP:conf/soda/ArthurV07}.
The beautiful idea of this approach is to choose additional seeds with the probability proportional
to their distance to the nearest seed (the first seed is picked uniformly).
If we assume there is a
cluster of points and no seed nearby, the probability mass of this cluster is substantial, and we are
likely to place the next seed there; afterwards the probability mass of this cluster reduces.
Furthermore, this initialization is (in expectation) $O(\log k)$ competitive to the optimal solution,
so it will theoretically generate good starting conditions.
But as seen in our experiments, this guarantee is pretty loose; and BUILD empirically
produces much better starting conditions than $k$-means++
(we are not aware of a detailed theoretical analysis).
But it is easy to see that in BUILD each medoid is chosen as a current optimum with respect to \TD{};
whereas $k$-means++ picks the first point randomly, and subsequent points are
(in expectation) random points from different clusters, but $k$-means++ makes no effort to find the medoid
(which is not that important for seeding $k$-means, where the mean is likely in between data points anyway).
Therefore, we must expect that with $k$-means++ we need around $k$ swaps just to pick the
medoid of each cluster (and hence, $k$ iterations of original PAM SWAP, although much fewer
with FastPAM2, experimentally confirmed later in \refsec{sec:numiter}).
Because a single iteration of swap used to take as much time as BUILD,
the $k$-means++ initialization only begins to shine if we use
FastPAM1 to reduce the cost of iterating together with the
FastPAM2 strategy of doing as many swaps as possible in each iteration.
\citet{DBLP:journals/datamine/LijffijtPP15} previously proposed to use k-means++ for PAM and CLARA;
but their complexity analysis misses a factor of $k$ for SWAP, and their experiments also only
used small $k$. Our experiments (in \refsec{sec:numiter}) show that
k-means++ initialization takes \emph{many more iterations} to converge
than with the original BUILD initialization;
so without the improvements introduced in this
article, it is usually not beneficial to use k-means++ with the original SWAP algorithm
for speed
(the benefit of randomness, the ability to get different results, remains;
and so do the theoretical guarantees).

\begin{algorithm2e}[t]
\caption{FastPAM LAB: Linear Approximate BUILD initialization.}
\label{alg:lab}
$(\TD,m_1) \leftarrow (\infty, \text{null})$\;
$S\leftarrow$ subsample of size $10+\lceil\sqrt{n}\rceil$ from $X$ \tcp*{Subsample}
\ForEach(\tcp*[f]{First medoid}){$x_j\in S$}{
  $\TD_j\leftarrow 0$ \;
  \lForEach{$x_o \in S \wedge x_o \neq x_j$}{
    $\TD_j \leftarrow \TD_j + \dist(x_o,x_j)$%
  }
  \lIf(\tcp*[f]{Smallest distance sum}){$\TD_j<\TD$}{
    $(\TD, m_1) \leftarrow (\TD_j, x_j)$%
  }
}
\For(\tcp*[f]{Other medoids}){$i=1\ldots k-1$}{
  $(\best{\Change}, \best{x}) \leftarrow (\infty, \text{null})$\;
  $S\leftarrow$ subsample of size $10+\lceil\sqrt{n}\rceil$ from $X\setminus\{m_1,\ldots,m_i\}$  \tcp*{Subsample}
  \ForEach{$x_j \in S$}{
    $\Change\leftarrow 0$ \;
    \ForEach{$x_o \in S \wedge x_o\neq x_j$}{
      $\delta \leftarrow \dist(x_o,x_j) - \min_{o \in m_1,\ldots,m_i}\dist(x_o, o)$\;
      \label{line:build-cached}
      \lIf{$\delta<0$}{$\Change\leftarrow\Change + \delta$}
    }
    \lIf(\tcp*[f]{best reduction in TD}){$\Change<\best{\Change}$}{
      $(\best{\Change}, \best{x}) \leftarrow (\Change, x_j)$%
    }
  }
  $(\TD, m_{i+1}) \leftarrow (\TD + \best{\Change}, \best{x})$\;
}
\Return $\TD, \{m_1,\ldots,m_k\}$\;
\end{algorithm2e}

We originally experimented with k-means++ initialization, but eventually settled
for a different strategy we call LAB (Linear Approximative BUILD).
What we title ``FastPAM'' then is the combination of LAB with the optimizations of FastPAM2.
As the name indicates, LAB is a linear approximation of the original PAM BUILD
(c.f., \refalg{alg:build}). In order to achieve linear runtime in $n$,
we simply subsample the data set. Before choosing each medoid, we sample
$10+\lceil\sqrt{n}\rceil$ points from all non-medoid points. From this subsample
we choose the one with the largest decrease $\Change$ with respect to the
current subsample only.
Results were slightly better with sampling $k$ times, and not just once;
since each object has $k$ chances of being in the sample, and if we draw a
bad sample it only affects a single medoid.
A pseudocode of LAB is given as \refalg{alg:lab}.

Clearly, this algorithm reduces the runtime of BUILD to $O(kn)$.
In the experiments in \refsec{sec:numiter},
the results with LAB were significantly better than with $k$-means++.

\subsection{Integration in CLARA: FastCLARA}

Since CLARA~(\citealt{KaufmanR86}, \citeyear[][Ch.{} 3]{KauRou90})
uses PAM as a subroutine, we can trivially
use our improved FastPAM with CLARA. In the experiments
(and the implementations provided as open-source) we
denote this variant as FastCLARA.

\subsection{Wider Exploration in CLARANS: FastCLARANS}\label{sec:better-clarans}

CLARANS~\citep{NgHan02} uses a randomized search instead of considering all possible swaps.
For this, it chooses a random pair of a non-medoid object and a medoid, computes whether this
improves the current loss, and then greedily performs this swap.
Adapting the idea from FastPAM1 to the random exploration approach of CLARANS,
we pick only the non-medoid object at random,
but consider all medoids for swapping at a similar cost to looking at a single medoid.
This means we can either explore $k$ times as many edges of the graph,
or we can reduce the number of samples to draw by a factor of $k$.
In our experiments we opted for the second choice, to make the results
comparable to the original CLARANS in the number of edges considered;
but as the edges chosen involve the same non-medoids, we expect a slight loss in quality
that should be easily countered by increasing the subsampling rate of non-medoids.
By varying the subsampling rate parameter, the user can easily control the tradeoff
between computation time and exploration.

\begin{figure*}[t]
\begin{subfigure}{.48\linewidth}
\includegraphics[width=\linewidth]{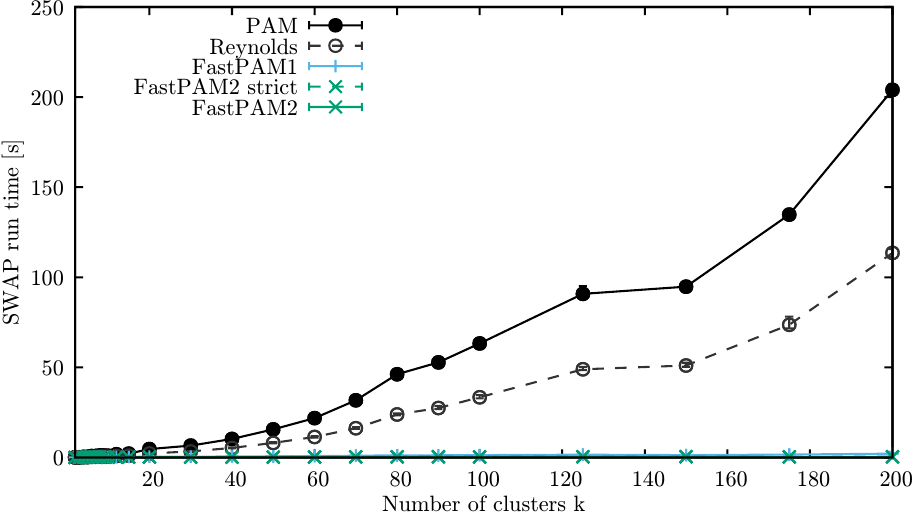}
\caption{Run time in linear space [ELKI]}
\label{fig:100plants-opttime-lin}
\end{subfigure}
\hfill
\begin{subfigure}{.48\linewidth}
\includegraphics[width=\linewidth]{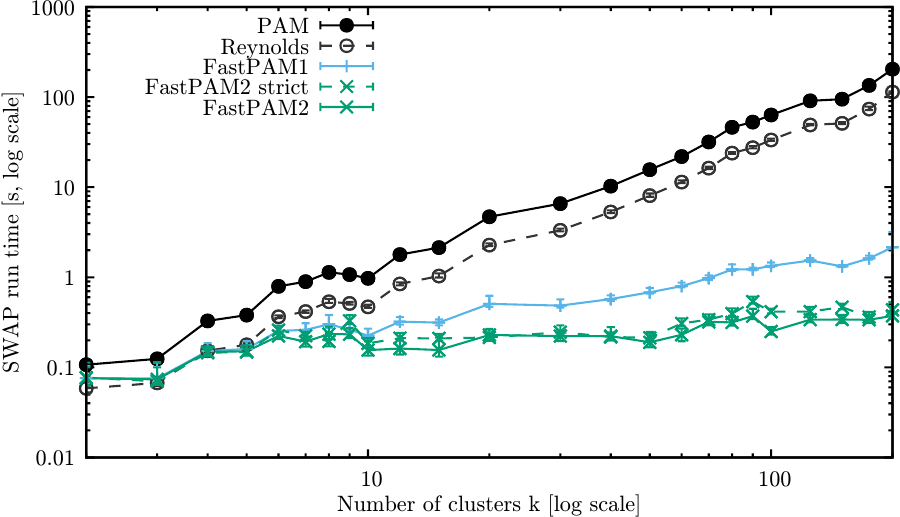}
\caption{Run time in log-log space [ELKI]}
\label{fig:100plants-opttime-log}
\end{subfigure}
\\
\begin{subfigure}{.48\linewidth}
\includegraphics[width=\linewidth]{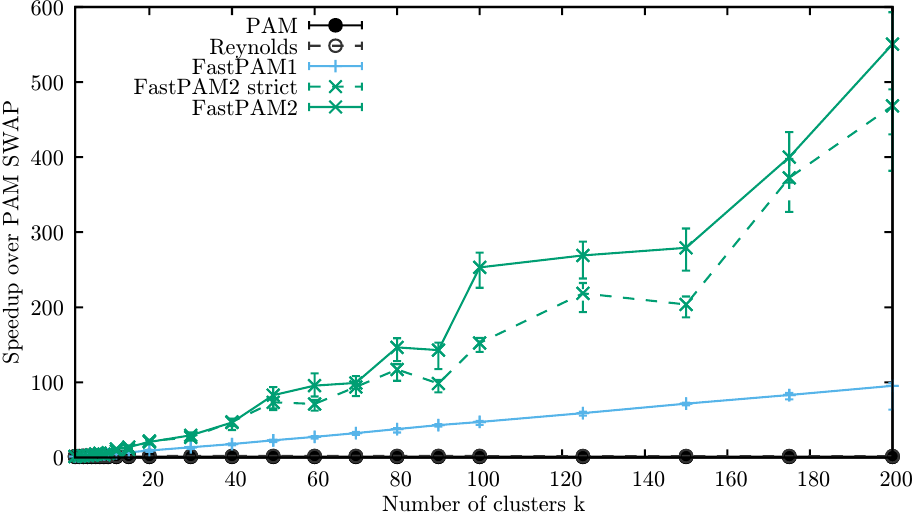}
\caption{Speedup in linear space [ELKI]}
\label{fig:100plants-optspeedup-lin}
\end{subfigure}
\hfill
\begin{subfigure}{.48\linewidth}
\includegraphics[width=\linewidth]{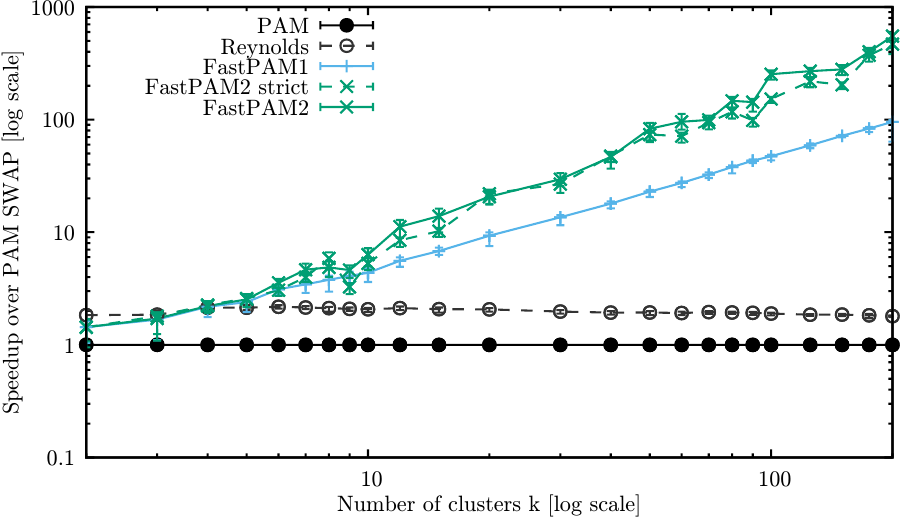}
\caption{Speedup in log-log space [ELKI]}
\label{fig:100plants-optspeedup-log}
\end{subfigure}
\\
\begin{subfigure}{.48\linewidth}
\includegraphics[width=\linewidth]{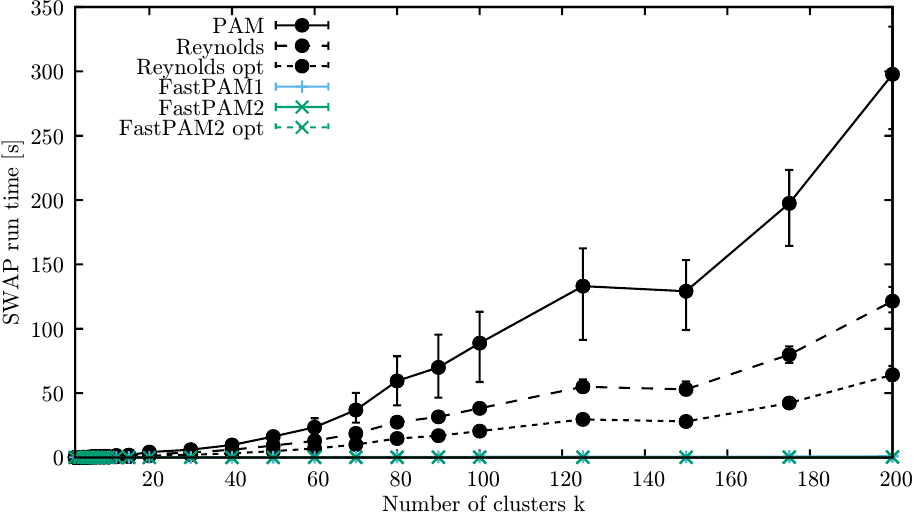}
\caption{Run time in linear space [R]}
\label{fig:100plants-opttime-lin-R}
\end{subfigure}
\hfill
\begin{subfigure}{.48\linewidth}
\includegraphics[width=\linewidth]{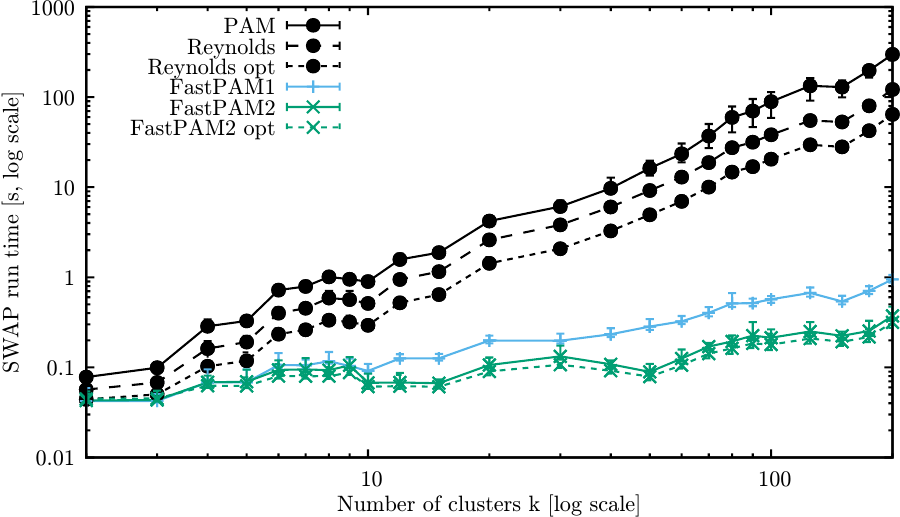}
\caption{Run time in log-log space [R]}
\label{fig:100plants-opttime-log-R}
\end{subfigure}
\\
\begin{subfigure}{.48\linewidth}
\includegraphics[width=\linewidth]{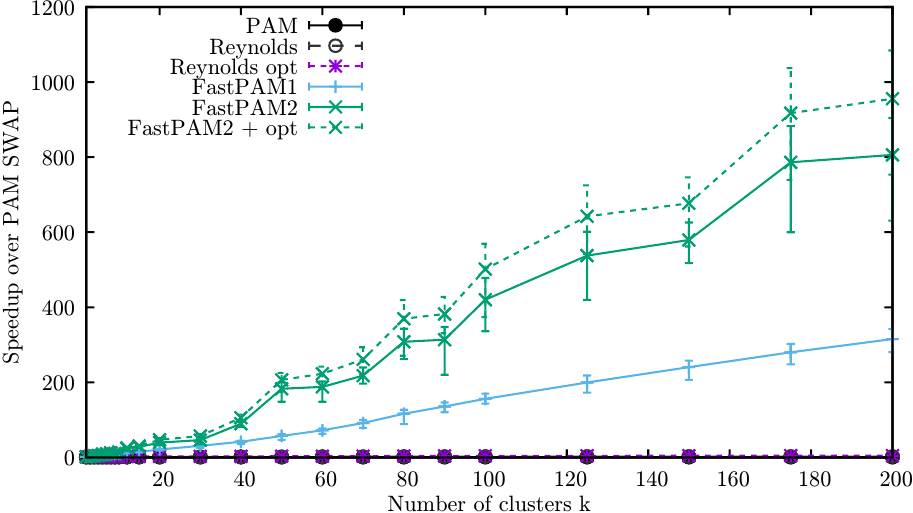}
\caption{Speedup in linear space [R]}
\label{fig:100plants-optspeedup-lin-R}
\end{subfigure}
\hfill
\begin{subfigure}{.48\linewidth}
\includegraphics[width=\linewidth]{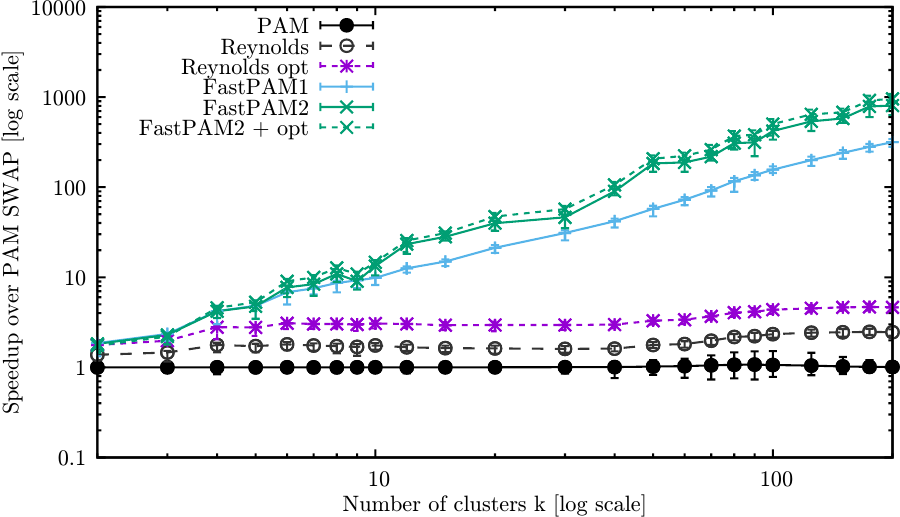}
\caption{Speedup in log-log space [R]}
\label{fig:100plants-optspeedup-log-R}
\end{subfigure}
\caption{Run time comparison of PAM SWAP (SWAP only, without DAISY, without BUILD)}
\label{fig:100plants}
\end{figure*}

\section{Experiments}

Theoretical considerations show that we must expect an $O(k)$ speedup
of FastPAM1 over the original PAM algorithm, so our experiments primarily need to
verify that there is no trivial error (in contrast to much work published
in recent years, the speedup is not just empirical).
Nevertheless constant factors and implementation details can make a
big difference~\citep{DBLP:journals/kais/KriegelSZ17}, and we want to ensure that
we do not pay big overheads for theoretical gains that would only manifest
for infinite data.\footnote{Clearly, our $O(k)$ fold speedup must be immediately
measurable, not just asymptotically, because the constant overhead for maintaining
the fixed array cache is small.}
Because of constant factors,
it could for example be possible that we need a certain minimum $k$ for this approach to be
beneficial over the original PAM.
For FastPAM2 we do not have such a theoretical argument for an additional speedup
over FastPAM1; and the speedup is expected to be a small factor due to the reduction
in iterations necessary.
In contrast to FastPAM1, it does not guarantee the exact same results; therefore we
also want to verify that they are of equivalent quality.
LAB, the third component for FastPAM, yields worse starting conditions. These should
not affect the final result much, but will require additional iterations of SWAP.
We observed increased runtimes when using $k$-means++ for PAM initialization, therefore it
needs to be verified experimentally that LAB does not require excessive additional iterations.

As primary data set for our experiments,
we use the ``One-hundred plant species leaves'' data set
(texture features only) from the well-known UCI repository \citep{Dua:2017}.
We chose this data set because it has 100 classes,
and 1600 instances, a fairly small size that PAM can still easily handle.
Naively, one would expect that $k=100$ is a good choice on this data set,
but some leaf species are likely not distinguishable by unsupervised learning.
We used the ELKI open-source data mining toolkit~\citep{DBLP:journals/pvldb/SchubertKEZSZ15}
in Java to develop our version. For comparison, we also ported FastPAM2 to the R \texttt{cluster}
package, which is based on the original PAM source code and written in C.
The source codes of both versions will be contributed to these projects.
Experiments were run on an Intel i7-7700 at 3.6 GHz with turbo boost disabled.
We perform 25 runs, and plot the average, minimum and maximum.
Both implementations show similar behavior, so we are confident that the results
are not just due to implementation differences~\citep{DBLP:journals/kais/KriegelSZ17},
and we verify the results on additional data sets.

\subsection{Run Time Speedup}
\begin{figure*}[t]
\begin{subfigure}{.48\linewidth}
\includegraphics[width=\linewidth]{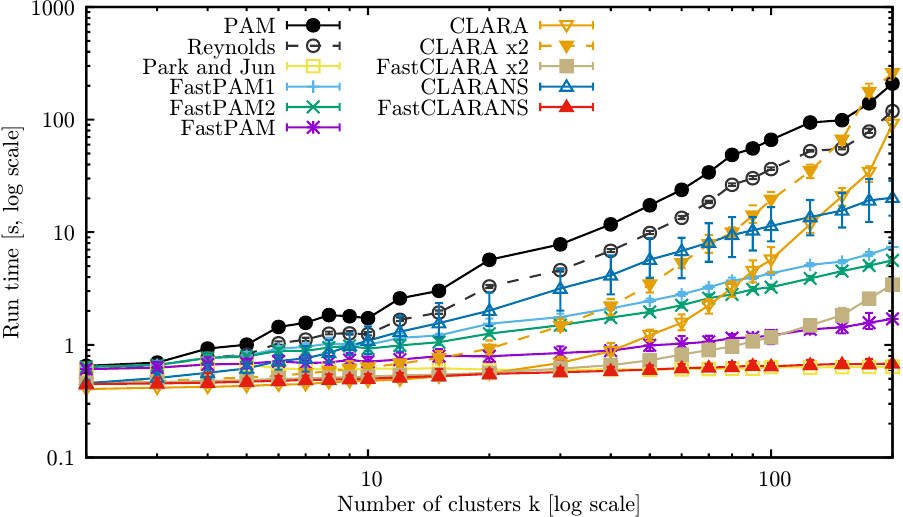}
\caption{Run time in log-log space [ELKI]}
\label{fig:100plants-runtime-log}
\end{subfigure}
\hfill
\begin{subfigure}{.48\linewidth}
\includegraphics[width=\linewidth]{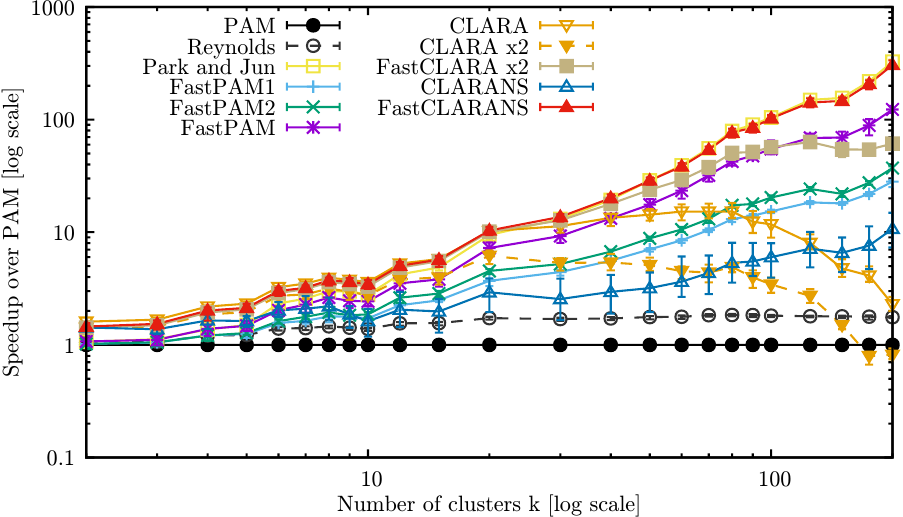}
\caption{Speedup in log-log space [ELKI]}
\label{fig:100plants-runspeedup-log}
\end{subfigure}
\caption{Run time comparison of different variations and derived algorithms.}
\label{fig:100plants-time}
\end{figure*}

In \reffig{fig:100plants}, we vary $k$ from 2 to 200, and plot the run time
of the PAM SWAP phase \emph{only} (the cost of computing the distance matrix and
the BUILD phase is not included),
using the original PAM, the Reynolds version,
as well as the proposed improvements.
\reffig{fig:100plants-opttime-lin} shows the run time in linear space,
to visualize the drastic run time differences observed. Reynolds' was quite
consistently two times faster than the original PAM;
but our proposed methods
were faster by a factor that grows approximately linearly with the number of
clusters $k$. In log-log-space, \reffig{fig:100plants-opttime-log}, we
can differentiate the three variants studied. While the ``greedy'' variant is
slightly faster than the ``strict'' variant of FastPAM2, the difference between these
two is not very large compared to the main contribution of this paper.
(Because this plot only includes the SWAP phase, LAB is not used here,
all methods are initialized with BUILD.)

In \reffig{fig:100plants-optspeedup-lin} we plot the speedup over PAM.
Reynolds' SWAP clearly was about twice as fast as the original PAM.
The FastPAM1 improvement gives an empirical speedup factor of about $\frac12k$,
while the second improvement contributed an additional speedup of about
$2\text{-}2.5\times$ by reducing the number of iterations.
Because of the multiplicative effect of these savings, the linear plot in
\reffig{fig:100plants-optspeedup-lin} gives the false impression that this
second contribution yields the larger benefit.
The logspace plot in \reffig{fig:100plants-optspeedup-log} more accurately
reflects the contribution of the two factors,
resulting in a speedup of over 250 times at $k=150$; while at $k=2$ and $k=3$
the speedup was just $1.4\times$ respectively $1.75\times$, and less than
our implementation of Reynolds (in R, as seen in \reffig{fig:100plants-optspeedup-log-R},
the difference at $k=2,3$ is negligible; so this is probably only an implementation difference).
In \reffig{fig:100plants-opttime-lin-R} to \reffig{fig:100plants-optspeedup-log-R}
we provide the results using the R implementation, which clearly exhibit similar behavior.
We only implemented the ``greedy'' $\tau=0$ version in R; but we additionally include
versions of Reynolds (\texttt{pamonce=2} of the existing package) and of our approach that
optimize the traversal of the distance matrix. In the most extreme case tested, a speedup
of about $1000\times$ at $k=200$ is measured -- but since the speedup is expected to
depend on $O(k)$, the exact values are meaningless, furthermore, we excluded the distance matrix
computation and initialization in this first experiment.

In \reffig{fig:100plants-time}, we study the run time of approximations to PAM
(including the distance matrix computation and initialization time now).
We only present the log-log space plots, because of the extreme differences.

The run time of CLARA, as $k$ increases, approaches the run time of PAM.
This is expected, because the subsample size for CLARA is chosen as $40+2k$,
and necessary because the subsample size needs to be sufficiently larger than $k$
(the recommendation of \citealt{DBLP:journals/datamine/LijffijtPP15}
of always using 80 samples is inappropriate for larger $k$).
For CLARA x2 we also evaluate doubling this value to $80+4k$,
and we also double the number of restarts from 5 to 10. CLARA x2 is thus
expected to take 8 times longer than CLARA, but should give better results.
FastCLARA is CLARA using our FastPAM approach, and performs much better, but for
large $k$ also eventually becomes slower than FastPAM.
The run time of CLARANS on this data set (see later for CLARANS problems) is in between
the original PAM and CLARA, and with our optimizations FastCLARANS becomes the fastest
method tested (at similar quality to CLARANS, and with the same problems).
\citeauthor{DBLP:journals/eswa/ParkJ09}'s (\citeyear{DBLP:journals/eswa/ParkJ09})
approach is similarly fast to FastCLARANS for large $k$, but its quality is quite poor,
as we will see and discuss in \refsec{sec:quality}.

\subsection{Number of Iterations}\label{sec:numiter}

We are not aware of theoretical results on the number of iterations needed for PAM.
Based on results for $k$-means, we must assume that the worst case is superpolynomial
like $k$-means~\citep{DBLP:conf/compgeom/ArthurV06}, albeit in practice
a ``few'' iterations are usually enough.
Because of this, we are also interested in studying the number of iterations,
depending on the choice of $k$ and the initialization method.

\begin{figure*}[t]
\begin{subfigure}{.48\linewidth}
\includegraphics[width=\linewidth]{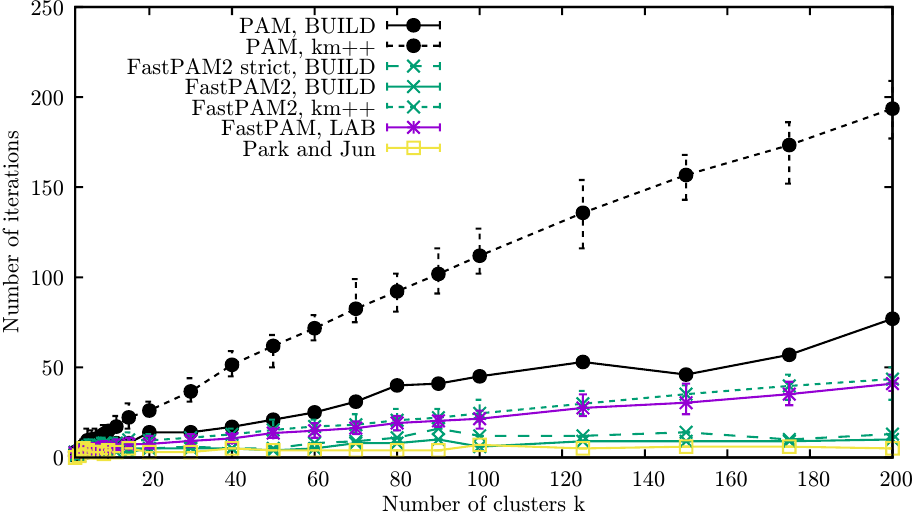}
\caption{Iterations in linear space [ELKI]}
\label{fig:100plants-iterations-lin}
\end{subfigure}
\hfill
\begin{subfigure}{.48\linewidth}
\includegraphics[width=\linewidth]{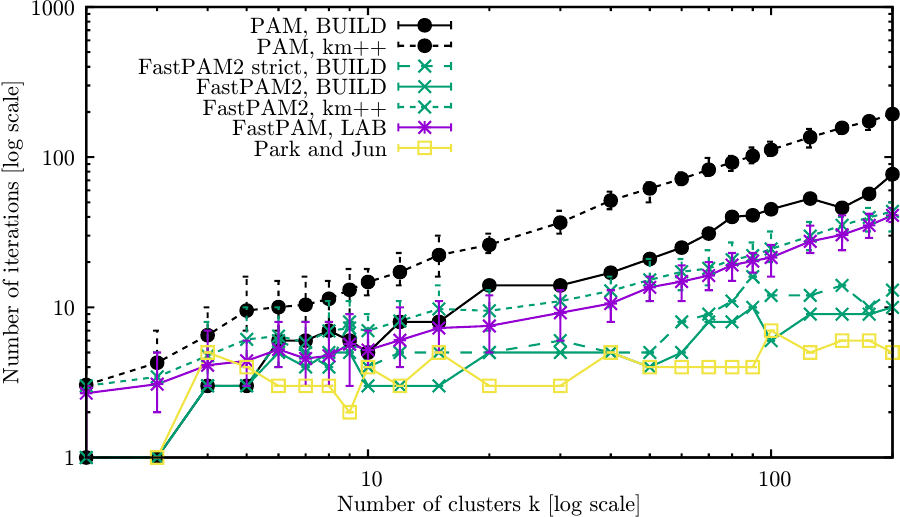}
\caption{Iterations in log-log space [ELKI]}
\label{fig:100plants-iterations-log}
\end{subfigure}
\caption{Number of iterations for PAM vs.{} FastPAM2 and BUILD vs.{} LAB initialization}
\label{fig:100plants-iter}
\end{figure*}

\begin{figure*}[t]
\begin{subfigure}{.48\linewidth}
\includegraphics[width=\linewidth]{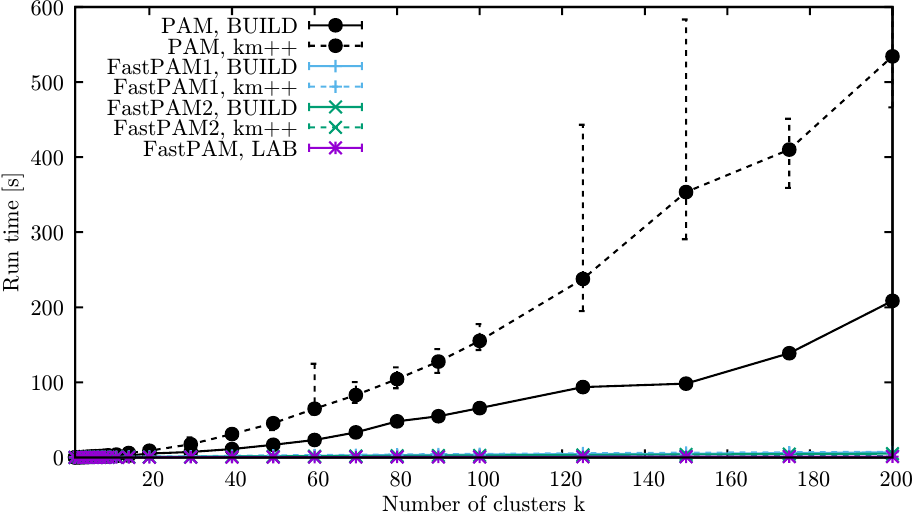}
\caption{Runtime with BUILD vs.{} $k$-means++ initialization in linear space [ELKI]}
\label{fig:100plants-iterations-lin}
\end{subfigure}
\hfill
\begin{subfigure}{.48\linewidth}
\includegraphics[width=\linewidth]{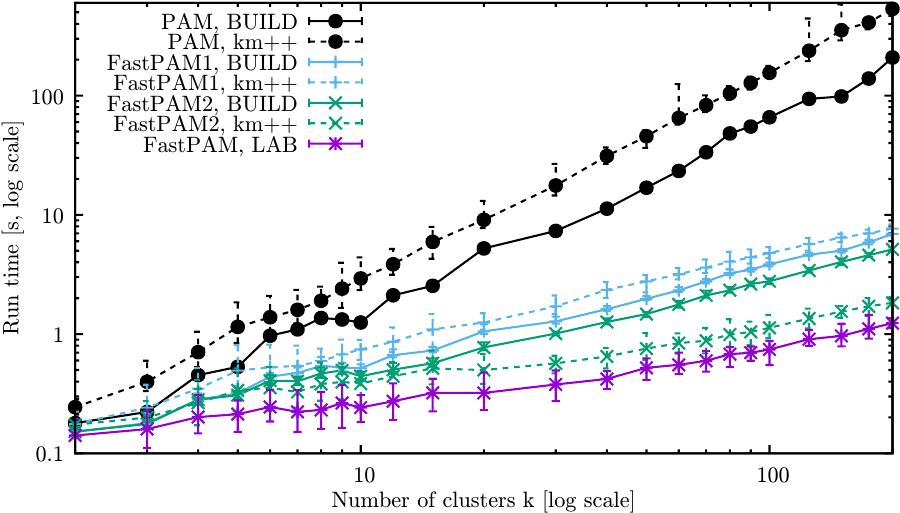}
\caption{Runtime with BUILD vs.{} $k$-means++ initialization in log-log space [ELKI]}
\label{fig:100plants-iterations-log}
\end{subfigure}
\caption{Runtime impact of $k$-means++ and LAB initialization}
\label{fig:100plants-runtimepp}
\end{figure*}

\reffig{fig:100plants-iter} shows the number of iterations needed with different methods,
both in linear space and log space.
In line with previous empirical results, only few iterations are necessary.
Because PAM only performs the best swap in each iteration, a linear dependency on $k$
is to be assumed; interestingly enough we usually observed much less than $k$ iterations,
so many medoids remain unchanged from their initial values (note that this may be due to
the rather small data set size, too). The \mbox{$k$-means++} initialization required roughly
2-4$\times$ as many iterations for PAM;
with the original algorithm where each iteration would cost
about as much as the BUILD initialization, this choice (although suggested by
\citealt{DBLP:journals/datamine/LijffijtPP15}) is detrimental even for small $k$.
With the improvements of this paper, these additional iterations are cheaper than the
rather slow BUILD initialization by a factor of $O(k)$ now, hence we can now begin with
a worse but cheaper starting point.
Furthermore, the FastPAM2 greedy approach which performs up to $k$ swaps in each iteration
does reduce the number of iterations substantially (the ``greedy'' version requires slightly
fewer iterations than the ``strict'' version, as expected). FastPAM2 with BUILD performed
the second-lowest numbers of iterations.
Our proposed LAB initialization of FastPAM saves a few extra iterations compared to the
$k$-means++ strategy, at better initial quality, and hence is measurably faster in the end.
\citet{DBLP:journals/eswa/ParkJ09} at first seems to perform very well in this figure,
with slightly fewer iterations than FastPAM2 with BUILD.
Unfortunately, this is because the ``$k$-means style'' algorithm misses many possible
improvements to the clustering, and hence produces much worse results as we will observe
in the next experiments.

In \reffig{fig:100plants-runtimepp} we revisit the runtime experiment, and focus on
initialization. As we can see, the increased number of iterations hurts
runtime with the original PAM algorithm as well as its Reynolds variant substantially
(the reasons for this are explained in \refsec{sec:fasterinit});
for FastPAM1, the use of $k$-means++ only comes at a small performance penalty
(while it still needs as many iterations as the original PAM, these have become $O(k)$ times
faster, and the initialization cost begins to matter much more),
and with FastPAM2's ability to perform multiple swaps per iteration,
a linear-time initialization such as $k$-means++ or the proposed LAB
clearly becomes the preferred initialization method, in particular for large $k$.

\subsection{Quality}\label{sec:quality}

Any algorithmic change and optimization comes at the risk of breaking some things,
or negatively affecting numerics (see, e.g., \citealt{DBLP:conf/ssdbm/SchubertG18} on
how common numerical issues are even with basic statistics such as variance in SQL databases).
In order to check for such issues, we made sure that our implementations pass the same
unit tests as the other algorithms in both ELKI and R. We do not expect numerical
problems, and Reynolds' variant and FastPAM1 are supposed to give the same result
(and do so in the experiments, so we exclude them from the plot).
The FastPAM2 algorithm is greedy in performing swaps, and may therefore converge to a
different solution, but that should be of the same quality, which we will verify now.

\begin{figure}[t]\centering
\includegraphics[width=.8\linewidth]{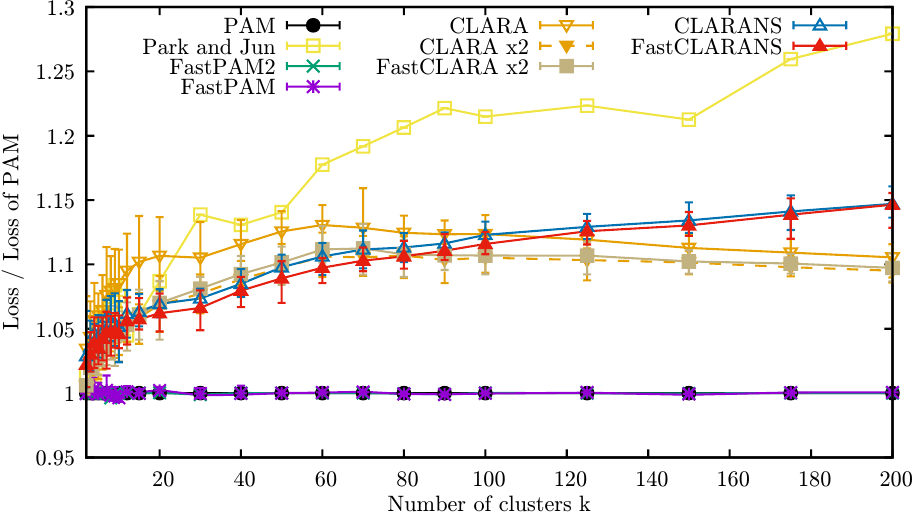}
\caption{Loss (\TD) compared to PAM}
\label{fig:100plants-cost}
\end{figure}

In \reffig{fig:100plants-cost} we visualize the loss, i.e., the objective \TD{}
of \refeqn{eqn:td},
of different approximations
compared to the solution found by the original PAM approach (which is not necessarily
the global minimum). For large $k$, the solution found by the approach of
\citet{DBLP:journals/eswa/ParkJ09} is over 25\% worse here, for the reasons discussed
before (worse initialization, misses many improvements).
Our strategy FastPAM2 even with the ``greedy'' approach gives results comparable to PAM
as expected (sometimes slightly better, sometimes slightly worse).
The cheaper LAB initialization (full FastPAM) does not cause a noticeable loss in quality
either, but further improves the total run time.
CLARA (which only uses a subsample of the data) finds considerably worse results.
By doubling the subsample size to $80+4k$ and doing twice as many restarts (CLARA x2)
the results only improve slightly for large $k$ (but much more for small $k$).
CLARA x2 is until about $k=70$ as good as CLARANS here, but faster; for larger $k$ it
becomes even better than CLARANS, but also slower. FastCLARA has the same quality as
CLARA x2 (we use the x2 parameters, too), but it was much faster.
FastCLARANS is slightly better than CLARANS, and was considerably faster.
All the CLARANS results degrade with increasing $k$,
so it may become necessary to increase the subsample size there, which will
increase the run time (it is up to the user to choose his preferences, quality or
runtime). In conclusion, all our ``Fast'' approaches perform as well as their older
counterparts, but are $O(k)$ times faster.

\begin{figure}[t]\centering
\includegraphics[width=.8\linewidth]{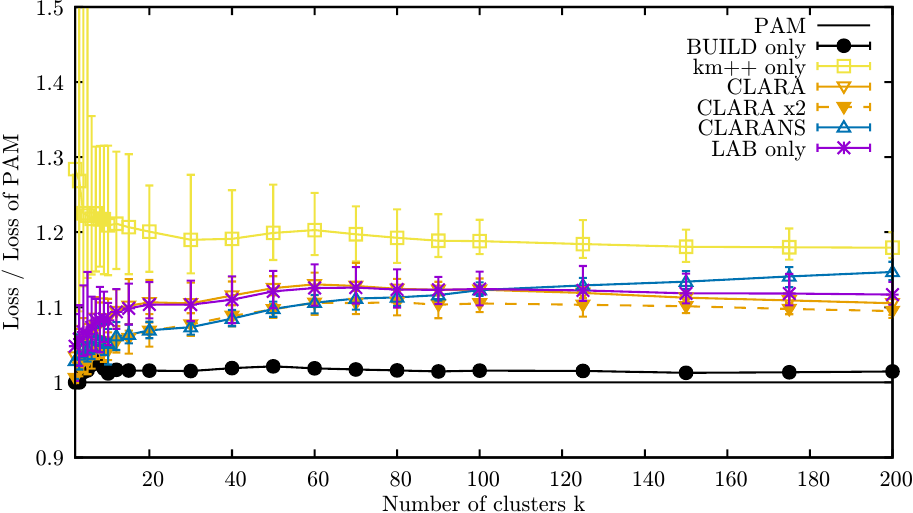}
\caption{Loss (TD) of $k$-means++ vs.{} BUILD initialization compared to PAM}
\label{fig:100plants-kmppinit}
\end{figure}

In \reffig{fig:100plants-kmppinit}, we evaluate the quality of LAB, $k$-means++,
and BUILD initialization compared to the converged PAM result. As seen in the previous
experiments, all three initializations will yield similar results after PAM, but we can
compare the quality of the initial medoids to the full PAM result.
As we can see, the BUILD approach produces the best initial results
(and as noted by \citealt[][Ch.{} 2]{KauRou87,KauRou90}, the BUILD result
may be usable without further refinement).
While $k$-means++ offers some theoretical advantages (c.f., \refsec{sec:fasterinit}),
the initial result is quite bad as this strategy only attempts to pick
a random point from each cluster, and not the medoids.
Our proposed LAB initialization is in between $k$-means++ and BUILD,
and by itself performs similar to CLARA. As it only considers a subset of the data,
its medoids will be worse than BUILD; but because it chooses the best medoid of the sample
it performs better than $k$-means++. Because it reduces the runtime for $O(n^2 k)$ to
$O(n k)$ it is the preferred choice for FastPAM nevertheless.

\subsection{Second Dataset}

\begin{figure}[t]
\begin{subfigure}{.49\linewidth}
\includegraphics[width=\linewidth]{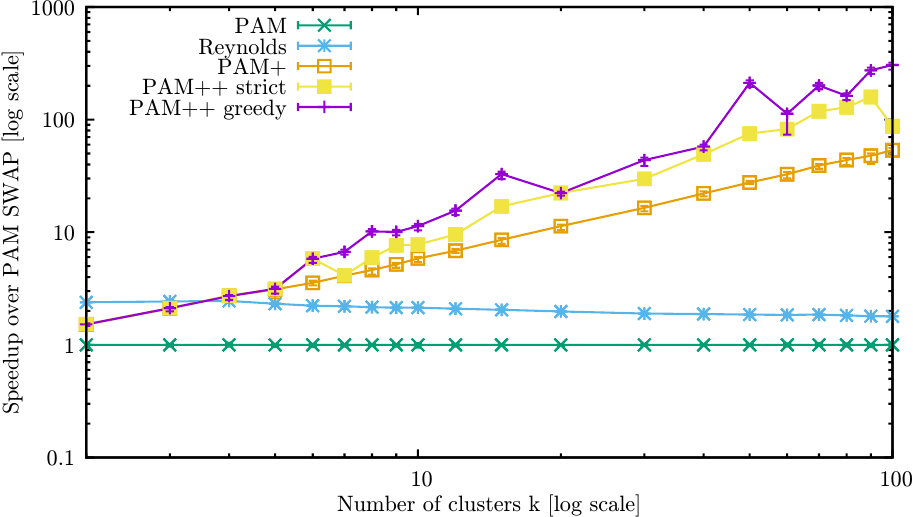}
\caption{SWAP Speedup [log-log]}
\label{fig:optdigits-optspeedup-log}
\end{subfigure}
\hfill
\begin{subfigure}{.49\linewidth}
\includegraphics[width=\linewidth]{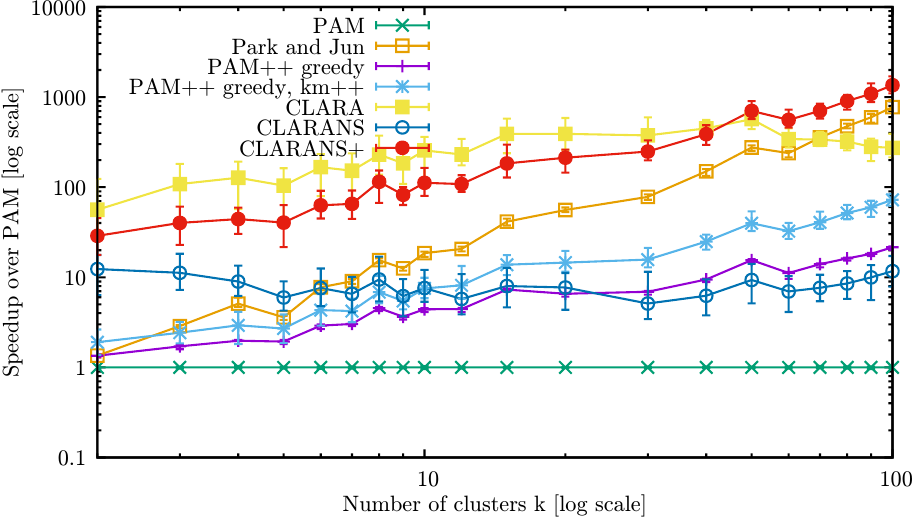}
\caption{Overall speedup [log-log]}
\label{fig:optdigits-runspeedup-log}
\end{subfigure}
\caption{Results on Optical Digits data using ELKI}
\label{fig:optdigits}
\end{figure}

We also verified our results on the ``Optical Recognition of Handwritten Digits''
data set from UCI~\citep{Dua:2017} with $n=5620$ instances, $d=64$ variables,
and $10$ natural classes.
An example of the results is shown in \reffig{fig:optdigits}
and also show an $O(k)$ speedup compared to PAM (we observe a $1000\times$ speedup
at $k=200$, but the more reasonable choice of $k$ would be $10$ here, where the speedup is
only about $10\times$). If a loss in quality is acceptable, FastCLARA and FastCLARANS again
are interesting alternatives outperforming their non-fast versions.
Clearly, the benefits on this data set are similar, and support our theoretical analysis.

\subsection{Scalability Experiments}

Just as PAM, our method also requires the entire distance matrix to be precomputed.
This will require $O(n^2)$ time and memory, making the method as-is unsuitable for big data
(for real big data problems, it will however often be good enough to cluster a sample that
still fits into memory -- for example with a mean, the precision only improves with $\sqrt{n}$,
so adding more data eventually does barely improve the results).
Our improvements focus on reducing the dependency on $k$, but we nevertheless experimented
with scalability in $n$, too (and we already included FastCLARA and FastCLARANS in the previous
experiments).
The behavior of the PAM variants is as expected $O(n^2)$, but we see nevertheless quite
big differences between PAM, FastPAM, and sampling-based approaches.
In this experiment, we use the well-known MNIST data set from the UCI repository \citep{Dua:2017},
which has 784 variables (each corresponding to a pixel in a $28\times28$ grid) and 60.000 instances.
We used the first $n=5000,10000,\ldots,35000$ instances with a time limit of 6 hours
and compare $k=10$ and $k=100$.
The high number of variables makes this data set expensive for CLARANS,
because it computes distances more than once.

\begin{figure}[t]
\begin{subfigure}{.49\linewidth}
\includegraphics[width=\linewidth]{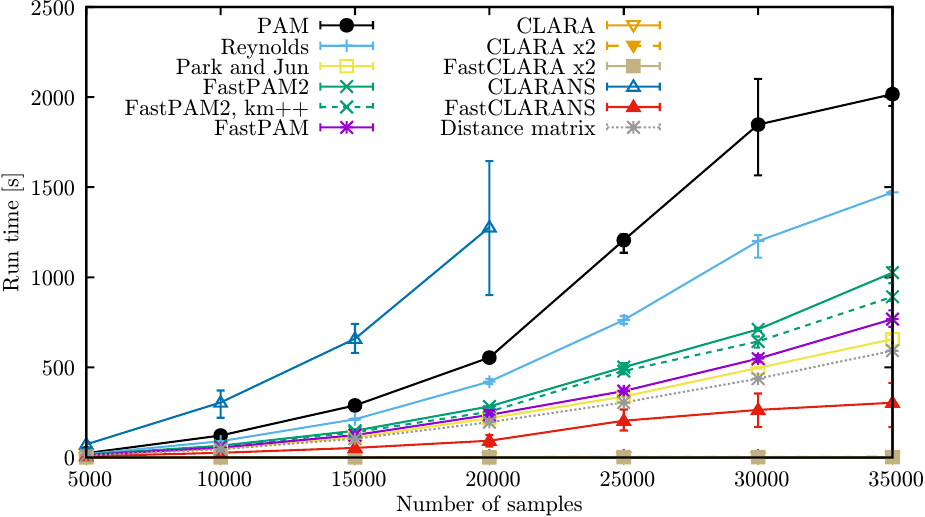}
\caption{Runtime with $k=10$}
\label{fig:mnist-runtime-10-lin}
\end{subfigure}
\hfill
\begin{subfigure}{.49\linewidth}
\includegraphics[width=\linewidth]{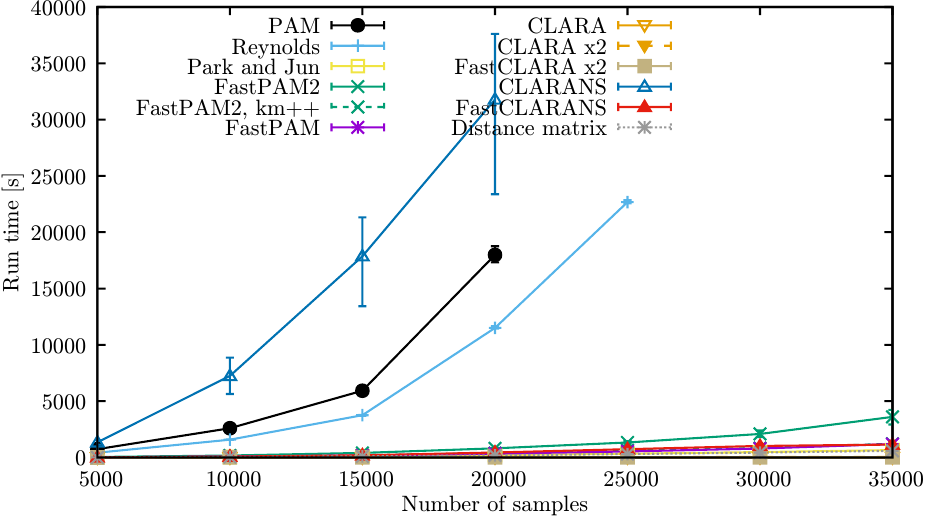}
\caption{Runtime with $k=100$}
\label{fig:mnist-runtime-100-lin}
\end{subfigure}
\\
\begin{subfigure}{.49\linewidth}
\includegraphics[width=\linewidth]{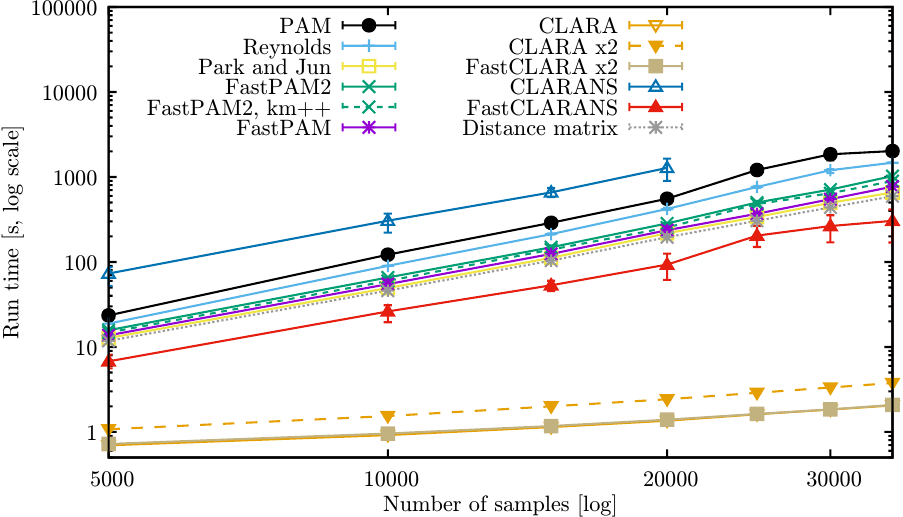}
\caption{Runtime with $k=10$ [log-log]}
\label{fig:mnist-runtime-10-log}
\end{subfigure}
\hfill
\begin{subfigure}{.49\linewidth}
\includegraphics[width=\linewidth]{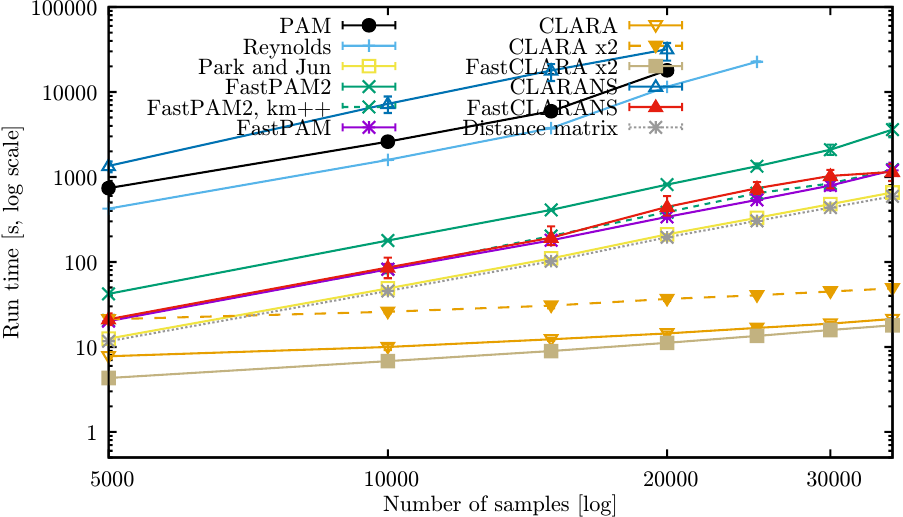}
\caption{Runtime with $k=100$ [log-log]}
\label{fig:mnist-runtime-100-log}
\end{subfigure}
\\
\begin{subfigure}{.49\linewidth}
\includegraphics[width=\linewidth]{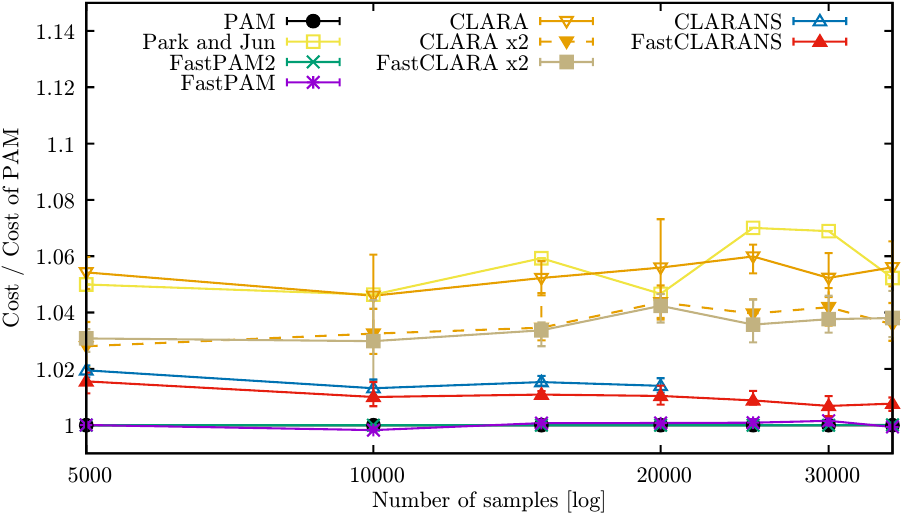}
\caption{Quality compared to PAM $k=10$}
\label{fig:mnist-costover-10}
\end{subfigure}
\hfill
\begin{subfigure}{.49\linewidth}
\includegraphics[width=\linewidth]{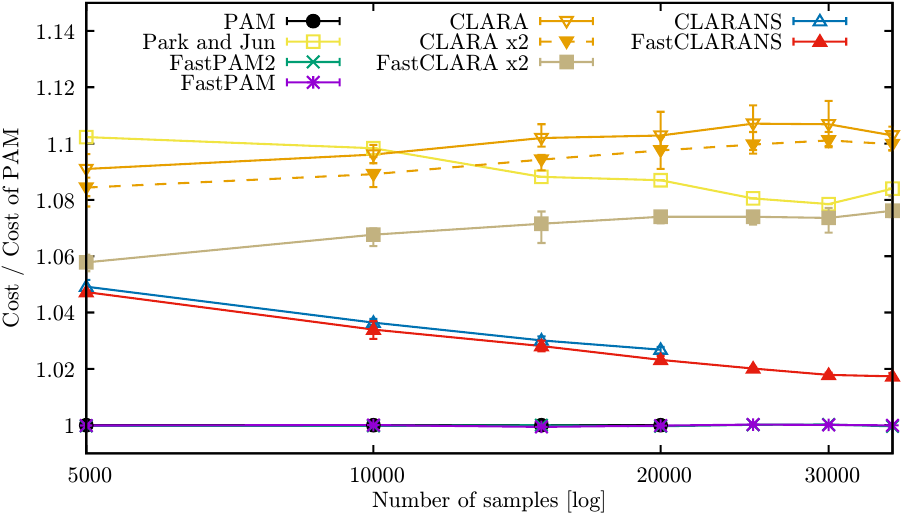}
\caption{Quality compared to PAM $k=100$}%
\label{fig:mnist-costover-100}
\end{subfigure}
\caption{Results on MNIST data using ELKI}
\label{fig:mnist-runtime}
\end{figure}

The problem of quadratic increase in runtime is best seen in the linear scale plots
\reffig{fig:mnist-runtime-10-lin} and \reffig{fig:mnist-runtime-100-lin}.
As a reference, we also give the time needed just for computing the distance matrix
as dotted line, which is also quadratic.
Except for CLARA and FastCLARANS, the runtime is dominated by computing the distance matrix
(and hence CLARA, which uses a constant-size sample independent of $n$, shines for large $n$).
The original CLARANS suffers from excessive distance re-computations. The authors assumed that
distances are cheap to compute, and noted that it may be necessary to cache the distances in
one way or another.
FastCLARANS reduces the number of distance computations of CLARANS by a factor
of $O(k)$, and is still cheaper than the full distance matrix here. For more expensive distances
such as dynamic time warping, FastPAM will outperform FastCLARANS, and it will almost always give
better results.
For $k=10$, only CLARANS, PAM and Reynolds' variant are problematic at this data size,
but at $k=100$ the benefits of our improvements become very noticeable.
The CLARA methods are squeezed to the axis in the linear plot, and hence we also provide
log-log plots in \reffig{fig:mnist-runtime-10-log} for $k=10$ and
\reffig{fig:mnist-runtime-100-log} for $k=100$.
For $k=10$, the lines of CLARA and FastCLARA x2 almost coincide by chance
(note that FastCLARA x2 produces a result comparable to the slower CLARA x2 method;
expected to be 8 times slower), but at $k=100$ it is faster than CLARA,
demonstrating that our improvements also accelerate CLARA by a factor of $O(k)$.

While the scalability in $n$ is quadratic as expected,
we observe that
if you can afford to compute the pairwise distance matrix, then you will \emph{now} also be
able to run FastPAM.
For $k=10$, the additional runtime of FastPAM was about 30\% the runtime of computing
the distance matrix computation,
and at $k=100$ FastPAM took about as much time as the distance matrix.
Hence, if you can compute the distance matrix, you can also run FastPAM for reasonable
values of $k\ll n$, and the main scalability problem will often be the memory consumption
of the distance matrix.
Without our optimizations, the cost of PAM would have been many times higher.

If computing the distance matrix is prohibitively expensive, it may still be possible to use
FastCLARA (CLARA with our improved FastPAM on the individual samples);
which is $O(k)$ times faster than original CLARA,
and will scale linearly in $n$. But as seen in \reffig{fig:mnist-costover-10} and
\reffig{fig:mnist-costover-100}, CLARA will usually give worse results
(about 10\% in our experiments). For many users this difference will be acceptable,
as a clustering result is never ``perfect''.
For large data sets, FastCLARANS will usually give better results,
unless the sample size of CLARA is increased considerably.
But on the other hand, FastCLARANS is only advisable for inexpensive distance functions such
as (low-dimensional) Euclidean distance, and would require using a non-trivial distance cache
for good performance otherwise.

\pagebreak

\section{Conclusions}

In this article we proposed a modification of the popular PAM algorithm
that typically yields an $O(k)$ fold speedup, by clever caching of partial
results in order to avoid recomputation. This caching was enabled by
changing the nesting order of the loops in the algorithm, showing once more
how much seemingly minor looking implementation details can
matter~\citep{DBLP:journals/kais/KriegelSZ17}.
As a second improvement, we propose to find the best swap for each medoid,
and execute as many as possible in each iteration,
which reduces the number of iterations needed for convergence without loss of quality,
as demonstrated in the experiments, and as supported by theoretical considerations.

The surprisingly large speedups obtained with this approach enable the use of
this classic clustering method on much larger data, in particular with
large $k$. Even such seemingly minor changes in such an algorithm can make a
big difference. It is hard to devise such things on the drawing board --
such solutions more naturally arise when trying to low-level optimize the code,
such as when and when not to allocate memory for buffers, and trying to avoid
recomputing the same values repeatedly. Today's compilers are reasonably good
at performing local optimization
(at least when it does not affect numerical precision, \citealt{DBLP:conf/ssdbm/SchubertG18}),
but will not introduce an additional array to cache such values.
With the faster refinement procedure, it now pays off to use cheaper
initialization methods with PAM.
We propose LAB initialization, a linear-time approximation of
the original PAM BUILD algorithm.

Methods based on PAM, such as CLARA, CLARANS, and the many parallel and distributed
variants of these algorithms for big data, all benefit from this improvement,
as they either use PAM as a subroutine (CLARA), or employ a similar swapping method (CLARANS)
that can be modified accordingly as seen in \refsec{sec:better-clarans}.

The proposed methods are included in the open-source
ELKI~\citep{DBLP:journals/pvldb/SchubertKEZSZ15} framework in version 0.7.5 \citep{DBLP:journals/corr/abs-1902-03616},
and FastPAM2 is included in the R \texttt{cluster} package version 2.0.9
(LAB, FastCLARA, and FastCLARANS are, however, not implemented for R yet, only in ELKI),
to make it easy for others to benefit from these improvements.
With the availability in two major clustering tools, we hope
that many users will find using PAM, CLARA, CLARANS, and later derived methods,
possible on much larger data sets with
higher $k$ than before.

\bibliographystyle{spbasic}
\bibliography{fasterpam.bib}

\end{document}